# Joint Application of the Target Trial Causal Framework and Machine Learning Modeling to Optimize Antibiotic Therapy: Use Case on Acute Bacterial Skin and Skin Structure Infections due to Methicillin-resistant *Staphylococcus aureus*

Inyoung Jun
Department of Epidemiology
University of Florida
Gainesville, FL, USA
inyoungjun@ufl.edu

Simone Marini
Department of Epidemiology
University of Florida
Gainesville, FL, USA
simone.marini@ufl.edu

Christina A. Boucher
Department of Computer and
Information Science and
Engineering
University of Florida
Gainesville, FL, USA
christinaboucher@ufl.edu

J. Glenn Morris
Emerging Pathogens Institute
University of Florida
Gainesville, FL, USA
jgmorris@epi.ufl.edu

Jiang Bian
Department of Health Outcomes &
Biomedical Informatics
University of Florida
Gainesville, FL, USA
bianjiang@ufl.edu

Mattia Prosperi[†]
Department of Epidemiology
University of Florida
Gainesville, FL, USA
m.prosperi@ufl.edu

## ABSTRACT

Bacterial infections are responsible for high mortality in the US and worldwide. Possible antimicrobial resistance underlying the infection, and multifaceted patient's clinical status can hamper the correct choice of antibiotic treatment. Randomized clinical trials provide average treatment effect estimates but are not ideal for risk stratification and optimization of therapeutic choice, i.e., individualized treatment effects (ITE) or counterfactuals. Here, we leverage large-scale, real-world electronic health record data, collected from Southern US academic clinics, to emulate a clinical trial, i.e., 'target trial', and develop a machine learning model of mortality prediction and ITE estimation for patients diagnosed with acute bacterial skin and skin structure infection (ABSSSI) due to methicillin-resistant *Staphylococcus* aureus (MRSA). ABSSSI-MRSA is a challenging condition with reduced treatment options—vancomycin is the preferred choice, but it has non-negligible side effects. First, we use propensity score matching to emulate the trial, and create a treatment-randomized (vancomycin vs. other antibiotic) dataset. Next, we use this data to train various machine learning methods (including boosted/LASSO logistic regression, support vector machines (SVM), and random forest (RF)) and choose the best-performing ones in terms of area under the receiver characteristic (AUC) through bootstrap validation. Lastly, we use the models to calculate ITE and identify possible averted deaths by therapy change. The out-of-bag tests indicate that SVM and RF are the most accurate, with AUC of 81% and 78%, respectively, but BLR/LASSO is not far behind (76%). By calculating the counterfactuals using the BLR/LASSO, vancomycin increases risk of death, but it shows large variation (odds ratio 1.2, 95% range 0.4-3.8) and the contribution to outcome probability is modest. Instead, the RF exhibits stronger changes in ITE, suggesting more complex treatment heterogeneity.

## 1 Introduction

Acute bacterial skin and skin structure infection (ABSSSI) includes various skin and skin structure infections, such as cellulitis, wound infections, and major abscesses. ABSSSIs are a common reason for emergency department visits and account for ~2% of hospital admissions in the United States [2,9], with a mortality of 15% (inpatient) to 8% (180-days) [12]. There have been continuous efforts to find the optimal antibiotic for ABSSSI to minimize complications and readmissions [21]. The Infectious Diseases Society of America guideline recommends therapeutic indications for each infection identified by clinical manifestations and severity [16]. As a treatment option for ABSSSI, vancomycin, linezolid, tigecycline, daptomycin, ceftaroline, and telavancin are suggested for severe infections; trimethoprim-sulfamethoxazole and doxycycline are suggested less severe conditions. Notwithstanding guideline recommendations, the choice of antibiotic for ABSSSI can vary greatly due to case presentation characteristics [2,22].

When ABSSSI is caused by methicillin-resistant *Staphylococcus aureus* (MRSA), vancomycin is one of the first recommended treatment options, however, vancomycin increases the risk of side effects, such as vancomycin-associated acute kidney injury [8].

While several studies suggest that other antibiotics (e.g., linezolid) could outperform vancomycin in terms of clinical success for patients with complicated forms of ABSSSI [6,7,23]. Randomized clinical trials are usually able to provide an unbiased, accurate estimate of the treatment effect with respect to a given outcome in a population of interest, however, these trials have difficulty obtaining heterogeneous study populations so they provide at best average treatment effect (ATE) estimates. Thus, clinical trials are not effective for risk stratification and optimization of therapeutic choice. Observational data like electronic health record (EHR) offer larger sample size and more variables than trials and allow the development of risk prediction models tailored to different patients' clinical statuses. Unfortunately, such observational data can contain multiple bias, e.g., in treatment assignment, and thus cannot be used for calculating outcomes of alternative therapy choices, commonly referred as 'counterfactuals' or individualized treatment effects (ITE), because such models are not guaranteed to be causal [18]. Causal inference helps with estimation of treatment effects from observational data, and its fusion with big data and artificial intelligence into 'causal AI' is opening great opportunities to develop models that accurately predict counterfactuals [1]. A review on counterfactual prediction in precision medicine was made by Lin *et al.* [11], highlighting scarcity of applied works. New studies are emerging, e.g., statin usage to prevent Alzheimer's disease [17].

Here, capitalizing on causal AI, we leverage real-world, large-scale longitudinal EHR data to cover the research gap about therapy optimization for ABSSSI-MRSA. More specifically, using data from 50,000+ patients seen in Southern US clinics over a decade, we emulate a clinical trial (vancomycin vs. other antibiotic) using the 'target trial' framework [3], and develop a machine learning model of mortality and counterfactual ITE prediction [1].

## 2 Methods

### 2.1 Data source, population settings, and trial emulation

We used a large academic hospital network EHR in Florida, the University of Florida (UF) Health's Integrated Data Repository (IDR). The study was approved as exempt by UF's institutional review board (protocol IRB201900652). Diagnostics and procedural EHR data from IDR are encoded through the World Health Organization's International Classification of Diseases (ICD) v.9, while for lab tests and medications are encoded through the Logical Observation Identifiers Names and Codes (LOINC) and RxNorm. The population screening criteria before the target trial eligibility procedure were to have had at least one antibiotic susceptibility test order (i.e., antibiogram) or a diagnosis of antibiotic-resistant infection at any time during the whole medical history. The target trial inclusion/exclusion criteria were defined on the basis of the clinical trial carried out by Itani *et al.* comparing vancomycin vs. linezolid in ABSSSI-MRSA [7], and are detailed in **Figure 1**. In brief, we included adult (≥18 years) patients admitted to inpatient/outpatient clinics and diagnosed ABSSSI-MRSA, with at least two symptoms among cellulitis / abscess,

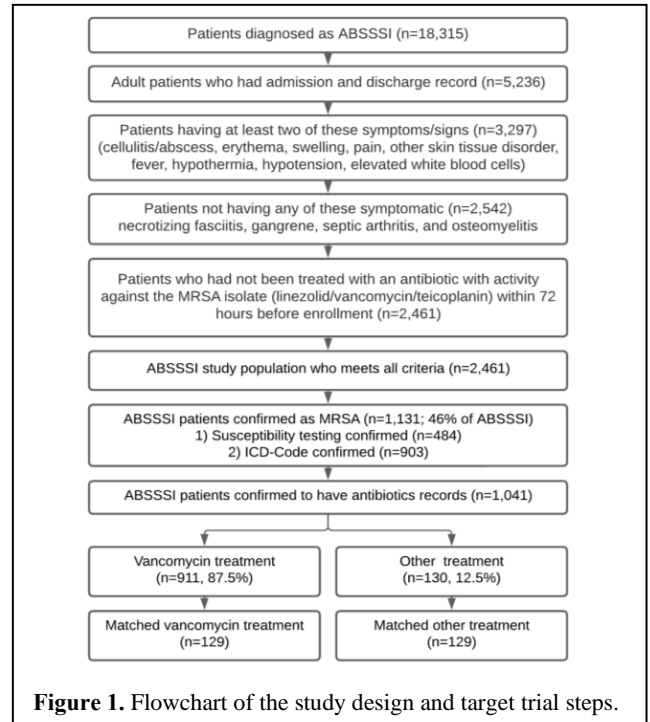

**Figure 1.** Flowchart of the study design and target trial steps.

erythema, swelling, pain, other skin/tissue disorder, fever, hypothermia, hypotension, and elevated blood cells. We used multiple ICD-9 codes validated from a previous study for identifying ABSSSI [10], including (but not limited to) 680.x, 681.x, 682.x, 684.x, 686.x, 704. For the symptoms, we also used relevant ICD-9 codes, including (but not limited to) 941.1, 780.6x, 288.6. Patients with any necrotizing fasciitis, gangrene, septic arthritis, osteomyelitis, were excluded. Patients needed not to have been treated with any antibiotic recommended for MRSA (linezolid, vancomycin, or teicoplanin) within 72 hours before admission. The ascertainment of MRSA infection was done by looking at antibiogram test result for methicillin, and *S. aureus* / MRSA ICD-9 codes, including 790.7, 995.91, 038.10, 041.12, 038.19. Of note, all ICD-10 codes were mapped to ICD-9 using R software package 'icd'. All patients eligible for the trial who received vancomycin upon admission constituted the treatment group, while those who received other antibiotic were the control group. The outcome/dependent variable was death within 90-days from admission. As covariates/predictors of the models to be inferred we included: demographics (Age, Sex, Race), medical conditions (body-mass index (BMI), diabetes, chronic kidney disease, human immunodeficiency virus infection (HIV), chronic obstructive pulmonary disorder (COPD), liver disease, cancer/malignancy, organ transplant, cardiovascular disease,

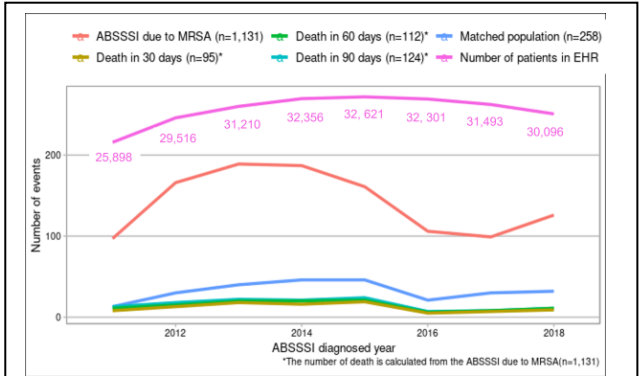

**Figure 2.** Incidence of ABSSSI-MRSA and of deaths in the study cohort (2011-2018).

Charlson's Comorbidity Index (CCI), C-reactive protein, prior evidence of antibiotic resistance, severity of illness (ICU stay, septic shock, qPitt score), and antibiotics treatment (vancomycin or other). We also included other ICD codes prior or concomitant to admission with at least a 10% frequency in the study population.

### 2.2 Propensity score matching

Formally, our problem is to evaluate the effectiveness of a treatment T (vs. control) on outcome Y among individuals with different background characteristics X. If the study was randomized, the T assignment would not be dependent on X and therefore the ATE would be 'easily' calculable as $E[Y|T=1] - E[Y|T=0]$. But in non-trial settings, the treatment T could be given on the basis of background characteristics (e.g., age, insurance, etc.) Thus, the effect of T on to Y could be biased by variables in X. One can identify a function $\pi(T=1|X)$, i.e., propensity to receive treatment T given X, and match all people with similar propensities (excluding those that do not), thus rebalancing data. This procedure is called propensity score matching (PSM) [20]. At this point, Y is independent on T given X, i.e. strong ignorability on treatment assignment (SITA) [5,14]. Then we can calculate ATE on the balanced population, but in precision medicine we are interested in the ITE $E[Y|T=t, X=x]$ and prediction of counterfactuals. For this study, we created a propensity score based on sex, age, race/ethnicity, BMI, CCI, qPitt score, and the ABSSSI symptoms at diagnosis. We used the software R package "matchit" [4], optimizing caliper width and nearest matching method.

### 2.3 Prediction and Counterfactual Modeling

We trained the following machine learning models on the PSM dataset: support vector machines (SVM), random forest (RF), LASSO and boosted logistic regression (BLR). All models were validated using bootstrapping (25 runs), calculating performance on out-of-bag samples. We used R software packages 'e1071', 'randomForest' and 'mboost'. SVM were tuned both with linear and radial basis kernels via grid search on parameters (gamma between 0.1 and 5, and cost between 0.01 and 10). For RF, number of trees was tested between 100 and 500 via out-of-bag error. For LASSO and BLR we used the package's default optimization routine. Performance functions included area under the curve precision-recall (AUCPR), sensitivity, specificity, and area under the receiver operating characteristic (AUC), with cut-offs optimized via the Euclidean approach [15]. After training and validation, we calculated the counterfactual probability of 90-days death by changing the therapy of each patient, e.g., if the patient did not take vancomycin, we put them in the treatment group, and vice versa. Then, using all factual and counterfactual predictions (i.e., ITE), we counted number of averted deaths.

## 3 Results

### 3.1 Data source, population settings, and trial emulation

The IDR data made available to us included EHR between 2011 and 2018 from total of 53,304 unique patients (a minimum of 25,898 in 2011 to a maximum of 32,621 in 2015). Among them, 50.9% were prescribed vancomycin at least once, while only 3.8% were ever prescribed linezolid. **Figure 2** illustrates trends in the incidence of ABSSSI-MRSA and the number of deaths among the total. Note that the total is not the general in-care population, as explained. Following the flowchart of **Figure 1**, the total number of patients with ABSSSI was 18,315, and then by applying all the target trial criteria, we obtained 1,041 individuals, of which 911 (87.5%) received vancomycin.

### 3.2 Propensity score matching

We then applied PSM, and the best caliper width (0.15) yielded a maximal sample size of 258 patients, with equal counts (129) in the vancomycin and non-vancomycin groups. The average standardized difference among variables in the two matched groups was 0.15. After matching, gender, CCI and qPitt were the most balanced between vancomycin vs. non-vancomycin groups, while BMI was still substantially different.

Through PSM, we obtained a dataset of 258 patients, with equal counts (129) in the vancomycin and non-vancomycin groups. The matching results of the study population are illustrated in **Figure 3**. We could observe that our matched population shared similar characteristics with the matched variables (all range within absolute standardized mean difference of 0.15). Details of before and after matching characteristics are illustrated in the **Supplementary Table 1**. Among the other diagnostic codes, 134 had a frequency >10%, and 102 were finally included as additional predictors based on a manual review. The top 3 frequent diagnoses were Shortness of breath (786.05), Long-term use of other medication (V58.69), and Anemia (285.9). Our main outcome – death in 90 days – was observed in 10.9% of vancomycin group and 9.3% of non-vancomycin group after matching, while it was 12.0% of vancomycin group and 9.2% of non-vancomycin group before matching.

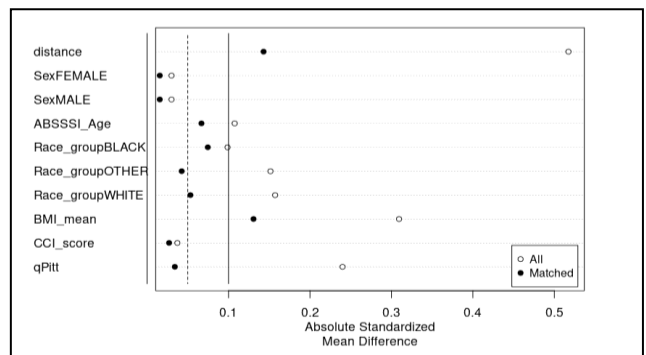

**Figure 3.** Difference of matched variables between each group in all datasets and matched dataset

### 3.3 Prediction and Counterfactual Modeling

On the PSM dataset, all statistical/machine learning methods showed similar prediction performance within 4% difference (**Table 1** and **Figure 4**). SVM (radial basis kernel) showed the best AUC overall, followed by RF, BLR, and LASSO.

**Table 1. Performance of prediction models from out-of-bag bootstrapping (n=25)**

| Model | AUC; Mean (SD) | AUCPR; Mean(SD) | Sensitivity | Specificity |
|---|---|---|---|---|
| LASSO | 0.760 (0.08) | 0.346 (0.072) | 0.764 | 0.736 |
| BLR | 0.764 (0.068) | 0.320 (0.073) | 0.818 | 0.716 |
| RF | 0.781 (0.058) | 0.311 (0.079) | 0.831 | 0.703 |
| SVM | 0.808 (0.053) | 0.383 (0.112) | 0.808 | 0.730 |

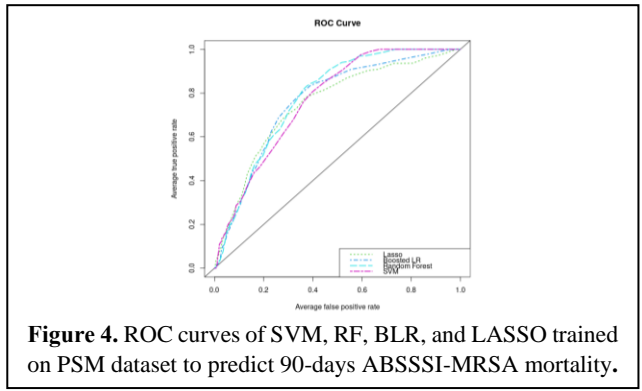

**Figure 4.** ROC curves of SVM, RF, BLR, and LASSO trained on PSM dataset to predict 90-days ABSSSI-MRSA mortality.

When looking at feature importance, in the BLR/LASSO vancomycin was almost never selected. By forcing it in the model, we obtained an odds ratio (OR) of 1.2 (95% range 0.4-3.8), indicating increased risk of death, but also large variation. The strongest predictor of death was Cancer/malignancy (OR = 9.45; CI 2.34 - 42.04). Other relevant features were Cardiomegaly (429.3), and pulmonary conditions (518.89, 793.19). The full list of LLR selected features is reported in **Supplementary Table 2**. When we looked at the random forest variable importance plot, we could observe that 276.0 (Hyperosmolality), shock, and the age at the ABSSSI are the top 3 most important risk factors in prediction in terms of mean decrease accuracy. When we evaluate based on the mean decease Gini, age at the ABSSSI, mean BMI, and 518.89 (Other lung disease) were selected as the top 3 most important predictors. (**Figure 5**). Finally, we evaluated the ITE and counterfactual predictions on the PSM dataset using both LASSO and RF (we chose these two models because one is linear in the coefficients and the other is nonlinear). In **Table 2**, we show how the risk of death would change if therapy was changed for a patient.

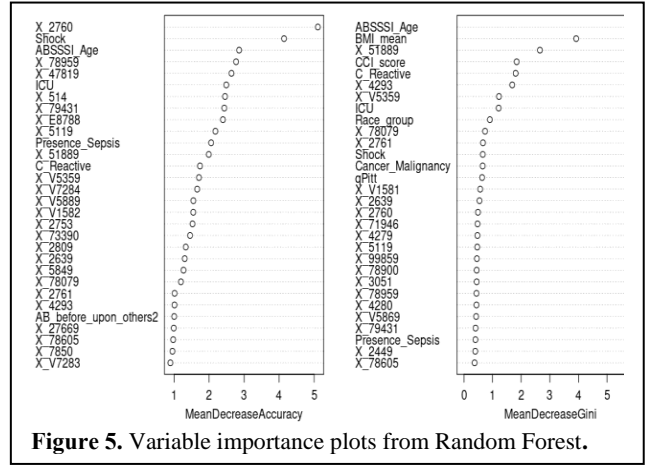

**Figure 5.** Variable importance plots from Random Forest.

In the LASSO, the risk of death always increases for patients switching from other antibiotic to vancomycin, since the OR is 1.2. Instead, the RF provides risk reduction for both groups, possibly due to interaction and covariates with the treatment variable.

**Table 2. Counts of patients at decreased/increased/ risk of death by switching therapy using the counterfactual prediction of ITE.**

| Model | Original Treatment | Reduced risk | Increased risk |
|---|---|---|---|
| LASSO | Vancomycin | 0 | 129 |
|  | Other antibiotic | 129 | 0 |
| RF | Vancomycin | 65 | 64 |
|  | Other antibiotic | 86 | 43 |

## 4  Discussion

In this work, we performed a target trial emulation in conjunction with machine learning to develop a counterfactual prediction model for optimization of ABSSSI-MRSA treatment. We found that nonlinear models (such as RF) are able to capture treatment heterogeneity and be useful in accurate ITE estimation. Conversely, in the (linear) logistic model, the ATE of vancomycin versus other antibiotics exhibited a large variance. The linear model can inform clinicians on the overall total effect of vancomycin - homogeneous among all patients. Instead, the RF allows more flexibility in looking for heterogeneous treatment effects. However, it is more difficult to interpret the RF model due to its inherent complexity. One drawback of our approach is reduced sample size due to PSM. Alternative could be using inverse probability weighting [19] or obtaining more patient data, for example from multiple EHR providers. Finally, another important point is that we picked up the first course of treatment during the admission but there could be changes when an antibiogram is performed, or if the patient does not respond. This will require more investigation of the EHR data but it warrants further investigation through dynamic treatment effect analysis, such as g-estimation [13]. In conclusion, this study shows that our causal AI approach can provide models with high prediction accuracy and causality with respect to an intervention of interest, i.e., antibiotic treatment choice. Intervention modelling is pivotal for improving patient care outcomes. Nonetheless, the development of such models from observational data should be always cautiously done because of possible unmeasured bias, even when causal methods are used.

## ACKNOWLEDGMENTS
This work was supported in part by: NIH NIAID R01AI145552 and HHS-NIH R01AG076234.

# Supplementary

S. Table 1. Characteristics of the target trial population before and after propensity score matching.

| Characteristics* | Before matching | | After matching | | Δ** |
|---|---|---|---|---|---|
| | Vanco (n=911) | Non-Vanco (n=130) | Vanco (n=129) | Non-Vanco (n=129) | |
| **Demographics** | | | | | |
| Age Mean (SD) | 52.0 (17.7) | 50.0 (18.2) | 51.2 (18.3) | 50.0 (18.2) | -0.8 |
| Sex Male | 516 (56.6%) | 76 (58.5%) | 74 (57.4%) | 75 (58.1%) | 0.8%p |
| Race White | 678 (74.4%) | 88 (67.7%) | 90 (69.8%) | 87 (67.4%) | -4.6 %p |
| Race Non-White | 233 (25.6%) | 42 (32.4%) | 39 (30.3%) | 42 (32.6%) | 4.7 %p |
| **Medical conditions** | | | | | |
| BMI Mean(SD) | 27.2 (8.75) | 26.1 (6.88) | 27.2 (6.78) | 26.3 (6.51) | 0 |
| Diabetes | 241 (26.5%) | 23 (17.7%) | 35 (27.1%) | 22 (17.1%) | 0.6%p |
| Chronic Kidney | 166 (18.2%) | 23 (17.7%) | 25 (19.4%) | 23 (17.8%) | 1.2%p |
| HIV | 27 (3.0%) | 2 (1.5%) | 4 (3.1%) | 2 (1.6%) | 0.1%p |
| COPD | 257 (28.2%) | 30 (23.1%) | 37 (28.7%) | 30 (23.3%) | 0.5%p |
| Liver Disease | 157 (17.2%) | 9 (6.9%) | 25 (19.4%) | 9 (7.0%) | 2.2%p |
| Cancer | 103 (11.3%) | 17 (13.1%) | 17 (13.2%) | 17 (13.2%) | 1.9%p |
| Organ transplant | 53 (5.8%) | 13 (10.0%) | 11 (8.5%) | 13 (10.1%) | 2.7%p |
| CVD | 624 (68.5%) | 73 (56.2%) | 83 (64.3%) | 72 (55.8%) | -4.2%p |
| CCI | 3.00 [0, 15.0] | 3.00 [0, 12.0] | 3.00 [0, 14.0] | 3.00 [0, 12.0] | 0 |
| C-reactive protein | 49.7 (84.4) | 36.9 (82.4) | 55.1 (86.2) | 37.0 (82.7) | 5.4 |
| **Antimicrobial Resistance (before/upon the infection)** | | | | | |
| Beta-lactams | 127 (13.9%) | 18 (13.8%) | 21 (16.3%) | 17 (13.2%) | 2.4 %p |
| Fluoroquinolones | 79 (8.7%) | 14 (10.8%) | 11 (8.5%) | 13 (10.1%) | -0.2 %p |
| Tetracyclines | 7 (0.8%) | 2 (1.5%) | 1 (0.8%) | 2 (1.6%) | 0 %p |
| Others | 111 (12.2%) | 15 (11.5%) | 15 (11.6%) | 14 (10.9%) | -0.6 %p |
| **Severity of illness** | | | | | |
| ICU stay | 279 (30.6%) | 27 (20.8%) | 28 (21.7%) | 27 (20.9%) | -8.9 %p |
| Septic shock | 105 (11.5%) | 5 (3.8%) | 11 (8.5%) | 5 (3.9%) | -3%p |
| qPitt score Mean(SD) | 0.187 (0.467) | 0.0692 (0.255) | 0.0853 (0.331) | 0.0698 (0.256) | -0.1017 |
| **Outcome – mortality** | | | | | |
| Death in 30 days | 86 (9.4%) | 8 (6.2%) | 12 (9.3%) | 8 (6.2%) | -0.1 %p |
| Death in 90 days | 109 (12.0%) | 12 (9.2%) | 14 (10.9%) | 12 (9.3%) | -1.1 %p |

*Mean (SD) or Median [Min, Max] or Count (%) ** Change of values for continuous variables or percentage point change for discrete variables of vancomycin group. (Δ = After matching – Before matching)

S. Table 2. Variables with strongest effect size on 90-days death by ABSSSI-MRSA selected by the BLR trained on PSM dataset.

| Variable | OR | CI (2.5%) | CI (97.5%) | Description |
|---|---|---|---|---|
| Cancer/ Malignancy | 9.45 | 2.34 | 42.04 | Cancer/Malignancy |
| 4293 | 5.27 | 1.65 | 18.21 | Cardiomegaly |
| 51889 | 4.58 | 1.38 | 16.24 | Other diseases of the lung, not elsewhere classified |
| 79319 | 4.54 | 1.16 | 20.28 | Other nonspecific abnormal findings of lung field |
| ICU | 2 | 0.59 | 6.8 | ICU stay |
| 486 | 1.94 | 0.55 | 6.78 | Pneumonia, organism unspecified |
| 4279 | 1.36 | 0.41 | 4.35 | Cardiac dysrhythmia, unspecified |
| 71946 | 1.27 | 0.28 | 5.41 | Pain in the joint, lower leg |
| Vancomycin* | 1.23 | 0.4 | 3.86 | Treatment |
| 78959 | 1.07 | 0.26 | 4.02 | Other ascites |
| 30000 | 0.89 | 0.18 | 4.02 | Anxiety state, unspecified |
| V5865 | 0.34 | 0.04 | 1.87 | Long-term (current) use of steroids |
| 34590 | 0.33 | 0.02 | 2.27 | Epilepsy, unspecified, without mention of intractable epilepsy |
| 27651 | 0.31 | 0.05 | 1.38 | Dehydration |
| E8788 | 0.31 | 0.04 | 1.31 | Other specified surgical operations and procedures causing the abnormal patient reactions, or later complications, without mention of misadventure at the time of operation |
| V7284 | 0.07 | 0.01 | 0.46 | Pre-operative examination, unspecified |

* Vancomycin is the treatment variable, forced in the model.